\pdfoutput=1
\documentclass[runningheads]{llncs}
\usepackage{float}
\usepackage[T1]{fontenc}
\usepackage{lmodern}
\usepackage{graphicx}
\usepackage{amsmath}
\usepackage{booktabs}
\usepackage{multirow}
\usepackage{subcaption}
\usepackage{array}
\usepackage{tabularx}
\usepackage{placeins}
\usepackage{marvosym}
\begin{document}
\title{Beyond In-Distribution Metrics: A Systematic Out-of-Distribution Evaluation of Congenital Heart Disease Segmentation}
\titlerunning{Evaluating OOD Generalization in CHD Segmentation}

\author{
  Aniketh Vijesh\inst{1}\textsuperscript{*} \and
  Shrisharanyan Vasu\inst{1}\textsuperscript{*} \and
  Abhijit Ramesh\inst{1} \and
  Clare Pomeroy-Ward\inst{1} \and
  Harikrishnan Anil Maya\inst{2} \and
  Sarin Xavier\inst{2} \and
  Mahesh Kappanayil\inst{2} \and
  Gilad Gressel\inst{1}\textsuperscript{\dag}
  }
  
  \authorrunning{A. Vijesh et al.}
  
  \institute{
  Amrita Vishwa Vidyapeetham, India\\
  \email{gilad.gressel@am.amrita.edu}
  \and
  Amrita Institute of Medical Sciences and Research Centre,
  Amrita Vishwa Vidyapeetham, Kochi, Kerala, India
  }

  \maketitle
  
  \begingroup
\renewcommand{\thefootnote}{*}
\footnotetext{equal contribution.}
\renewcommand{\thefootnote}{\dag}
\footnotetext{Corresponding author.}
\endgroup

\begin{abstract}

Congenital heart disease (CHD) diagnosis and surgical planning often require patient-specific 3D anatomical models, but manual segmentation is labor-intensive, particularly in complex anatomies. Although deep-learning methods can automate this process, they are typically evaluated in-distribution, despite clinically relevant shifts in scanner, protocol, institution, population, and imaging modality. We present, to our knowledge, the first systematic evaluation of out-of-distribution (OOD) generalization in CHD segmentation, using ImageCHD as a held-out target cohort. We compare representative segmentation architectures under combined CT and CMR training, CT-only training, self-supervised pretraining, and limited target-domain adaptation. In-distribution performance proves to be a poor indicator of cross-cohort robustness: nnU-Net achieves the highest validation Dice (0.77) but falls to 0.51 on ImageCHD, while SwinUNETR generalizes substantially better, reaching 0.67 Dice. MAE and JEPA pretraining provide only modest additional benefit, suggesting that architecture contributes more to robustness than the tested pretraining strategies in this setting. When limited target-domain supervision is introduced, all SwinUNETR variants exceed 0.76 Dice with only 11 labeled ImageCHD cases. These findings demonstrate that conventional in-distribution evaluation can obscure clinically important generalization failures and support explicit cross-dataset testing as a key component of CHD segmentation evaluation.

\keywords{Congenital heart disease \and Medical image segmentation \and
Out-of-distribution generalization \and Multi-modal learning \and
Self-supervised learning \and Cardiac CT \and Cardiovascular MRI}
\end{abstract}

\section{Introduction}

Congenital heart disease (CHD) often requires patient-specific 3D anatomical understanding for diagnosis, surgical planning, and catheter-based treatment~\cite{pace}. Cardiac CT and cardiovascular MR provide the necessary volumetric imaging, but accurate segmentation of chambers and great vessels remains labor-intensive in complex CHD anatomies~\cite{goo2020advanced,byrne2016segmentation,xu2019accurate}. Deep-learning methods for automatic segmentation have advanced from U-Net-style models to graph-based, hybrid CNN-attention, transformer-based, and self-supervised approaches~\cite{imagechd,hybrid,cardiacseg,qayyum25}. 

Despite strong reported results, existing CHD segmentation methods are evaluated in-distribution, where training and test data share the same dataset, modality, institution, and acquisition setting. Clinical deployment involves shifts in scanner, protocol, institution, population, and modality. Prior clinical AI systems, including IBM Watson for Oncology and Google Health's diabetic-retinopathy screening tool, illustrate that performance in curated settings does not necessarily transfer across hospitals, workflows, countries, and patient populations~\cite{ross2017watson,beede2020human}. Medical image segmentation models are similarly known to degrade under distribution shift~\cite{torpmannhagen,sanner,vasiliuk}. CHD segmentation is especially vulnerable because CHD anatomy is heterogeneous and long-tailed: rare anatomies, unusual combinations of defects, scanner-specific characteristics, and institution-specific acquisition patterns may be poorly represented or absent in any single training cohort~\cite{imagechd,kong2024sdf4chd}. Yet existing CHD segmentation studies do not explicitly benchmark out-of-distribution (OOD) generalization.

A related open question concerns modality. Prior CHD work largely treats CT and CMR as separate settings, although both image the same underlying cardiac anatomy through different acquisition processes. Whether joint CT/CMR training improves OOD generalization remains untested. We therefore evaluate multi-modal CT/CMR training alongside architecture and pretraining strategy.

We conduct, to our knowledge, the first systematic study of OOD generalization for CHD segmentation. Using ImageCHD as a held-out target, we evaluate nnU-Net~\cite{nnunet2021}, CardiacSeg~\cite{cardiacseg}, the layer-based hybrid encoder--decoder of Zhu et al.~\cite{hybrid}, hereafter referred to as Zhu-Net for clarity, and SwinUNETR~\cite{hatamizadeh2022swinunetrswintransformers}, trained from scratch or with MAE-style~\cite{he2022mae} and JEPA-style~\cite{assran2023ijepa} pretraining. We compare combined CT+CMR training, CT-only training, and target-domain adaptation with progressively added ImageCHD labels.

\section{Methodology}

We evaluate OOD generalization for CHD segmentation by systematically varying training cohort composition. We compare representative segmentation methods under three settings: multi-modal CT+CMR training, CT-only training, and limited target-domain adaptation, while using ImageCHD as the primary held-out target domain.

\begin{figure}
\centering
\includegraphics[width=0.8\linewidth]{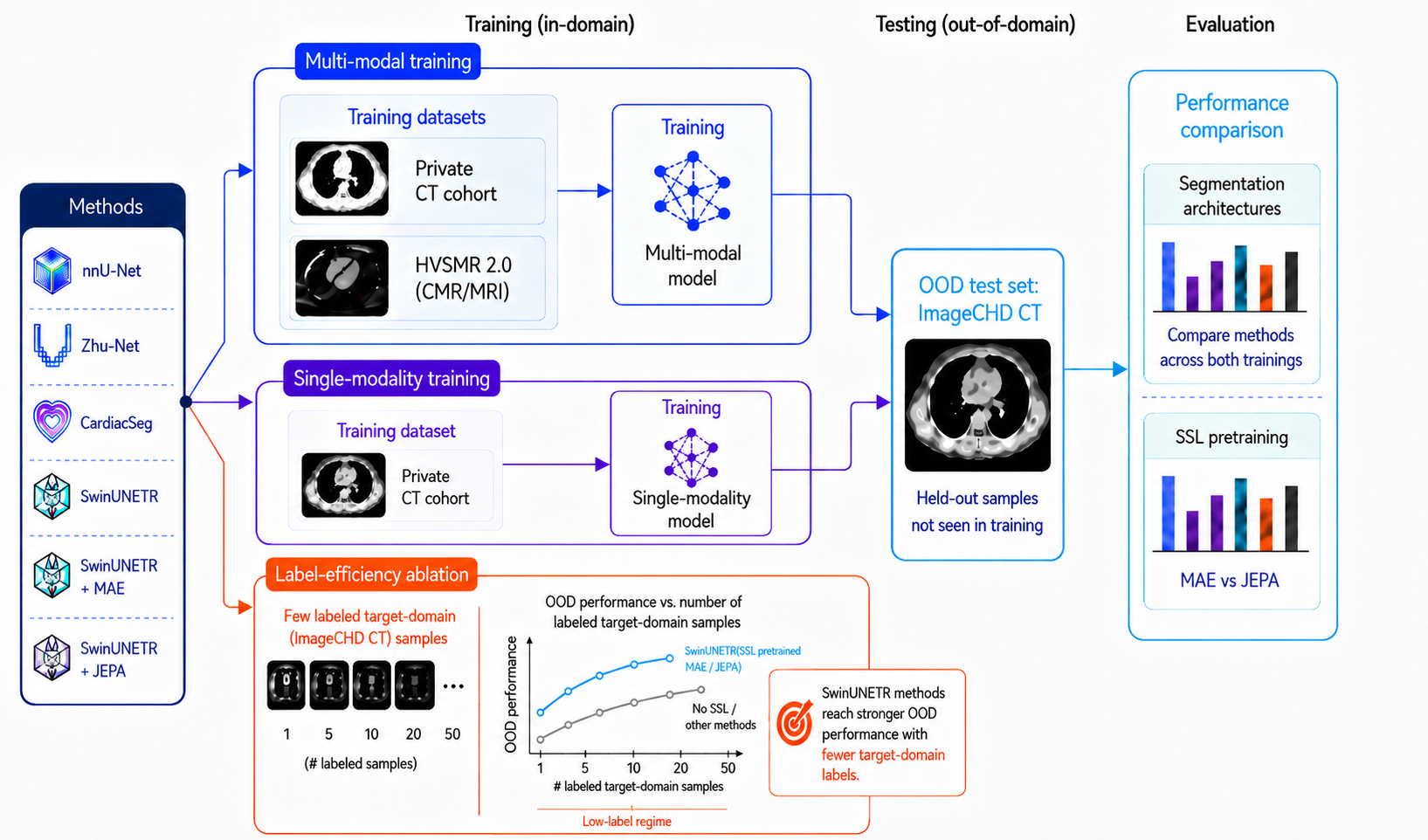}
\caption{Overview of the OOD evaluation design. Models are trained under CT+CMR, CT-only, and limited target-domain adaptation settings, then evaluated on held-out ImageCHD CT data using Dice, HD95, and ASSD.}
\label{fig:method_overview}
\end{figure}

Figure~\ref{fig:method_overview} summarizes the evaluation pipeline. Models are trained on either combined CT+CMR data, private CT data alone, or progressively expanded labeled ImageCHD subsets. Across these settings we compare four segmentation architectures and, for SwinUNETR, evaluate both training from scratch and encoder pretraining with Masked Auto-Encoding (MAE)~\cite{he2022mae} and Joint Embedding Predictive Architecture (JEPA)~\cite{assran2023ijepa}. Each model is fine-tuned end-to-end on labeled data, and performance is evaluated with Dice, HD95, and ASSD on both in-distribution validation data and held-out OOD data.

\subsection{Datasets for OOD Evaluation}
We use three patient-level 3D cardiac imaging cohorts to simulate clinically relevant shifts in modality, institution, scanner, and CHD case mix (Table~\ref{tab:datasets}). The training data consist of a private contrast-enhanced CT cohort, 3D-Labs, curated at Amrita Institute of Medical Sciences, Kochi, Kerala, India, and the public HVSMR-2.0 CMR cohort~\cite{hvsmr2}. ImageCHD, a public contrast-enhanced CT cohort, is reserved as the held-out OOD target~\cite{imagechd}. Across all cohorts, segmentation is evaluated on the same six anatomical structures: left ventricle, right ventricle, left atrium, right atrium, aorta, and pulmonary artery.

\begin{table}
\centering
\caption{Cohorts used for OOD evaluation. Counts show Train/Val/OOD splits for the multi-modal CT+CMR setting.}
\label{tab:datasets}
\small
\setlength{\tabcolsep}{5pt}
\renewcommand{\arraystretch}{1.15}
\begin{tabular}{@{}lllccl@{}}
\toprule
Cohort & Mod. & Source & $n$ & Train/Val/OOD & Dominant case-mix \\
\midrule
3D-Labs \textit{(priv.)} & CT  & Multi-vendor$^{a}$   & 79  & 71\,/\,8\,/\,--   & 16 phenotypes$^{b}$ \\
HVSMR-2.0                & CMR & Public               & 60  & 52\,/\,8\,/\,--   & Pre-/post-op CHD \\
ImageCHD                 & CT  & Public$^{c}$         & 110 & --\,/\,--\,/\,110 & Septal (VSD, ASD) \\
\bottomrule
\end{tabular}
\\[3pt]
{\footnotesize $^{a}$Amrita Institute of Medical Sciences, Kochi, Kerala, India; Philips, Siemens Healthineers, GE Healthcare.
\quad $^{b}$Most frequent: DORV, VSD, PS, CCTGA.
\quad $^{c}$Siemens Biograph 64.}
\end{table}

The 3D-Labs cohort provides a heterogeneous source domain, with multi-vendor acquisition and 16 CHD phenotypes spanning conotruncal defects, ventriculo-arterial discordance, outflow-tract obstruction, venous anomalies, and aortic arch abnormalities. Multi-defect anatomy is common, most often involving DORV, VSD, PS, and CCTGA. HVSMR-2.0 adds CMR cases with pre-operative and post-operative CHD anatomy, allowing us to test whether multi-modal CT+CMR training improves generalization beyond CT-only supervision.

We use ImageCHD as the primary OOD target because it is the most widely used public CHD segmentation dataset and differs from 3D-Labs in acquisition source, scanner setting, and diagnosis distribution. This design tests whether models trained on a heterogeneous private CT cohort, with or without additional CMR data, generalize to a distinct public CT cohort.

\subsection{Segmentation Methods}

We benchmark representative CHD segmentation methods spanning supervised encoder-decoder models, task-specific cardiac architectures, transformer-based segmentation, and self-supervised pretraining. The supervised models include nnU-Net~\cite{nnunet2021}, a strong self-configuring medical segmentation baseline, and Zhu-Net~\cite{hybrid}, a hybrid encoder-decoder model designed for CHD anatomy. We also evaluate SwinUNETR~\cite{hatamizadeh2022swinunetrswintransformers}, a U-shaped 3D segmentation network with a hierarchical Swin Transformer encoder and convolutional decoder, both trained from scratch and initialized with self-supervised pretraining.

For self-supervised models, we evaluate CardiacSeg~\cite{cardiacseg}, SwinUNETR with MAE pretraining, and SwinUNETR with JEPA pretraining. CardiacSeg is a CHD-specific masked-pretrained model combining a ViT encoder, scaling feature pyramid, and convolutional decoder. For SwinUNETR, MAE pretraining learns by reconstructing masked voxel intensities~\cite{he2022mae}, whereas JEPA predicts latent embeddings of masked regions using a context encoder, EMA target encoder, and predictor network~\cite{assran2023ijepa}. After pretraining, the auxiliary pretraining components are discarded and the pretrained encoder initializes the downstream segmentation model for end-to-end fine-tuning.

\noindent\textbf{Evaluation metrics and losses.}
Segmentation performance is evaluated using the Dice similarity coefficient (Dice), 95th-percentile Hausdorff distance (HD95), and average symmetric surface distance (ASSD). Dice measures volumetric overlap, while HD95 and ASSD measure boundary accuracy. HD95 and ASSD are reported in millimetres, with lower values indicating better boundary accuracy. Metrics are computed over foreground cardiac structures and reported on in-distribution validation and held-out OOD test cohorts. All supervised segmentation fine-tuning uses a Dice-based objective, with per-method details provided in Section~\ref{sec:implementation}. For self-supervised pretraining, MAE minimizes mean-squared error over masked voxel intensities, while JEPA minimizes mean-squared error in latent embedding space.

\section{Experiments and Results}

\subsection{Implementation Details} \label{sec:implementation}

All experiments use the cohort splits in Table~\ref{tab:datasets}. In the adaptation setting, labeled ImageCHD cases are added in increments of 11 and evaluated on a fixed 22-case holdout. Self-supervised models are pretrained only on the corresponding training pool, with no external data. Zhu-Net, CardiacSeg, and the SwinUNETR variants are implemented in PyTorch Lightning with MONAI and trained on a single NVIDIA RTX~6000 Ada GPU with 48~GB memory, using $128^3$ inputs and batch size~1.

\noindent\textbf{Preprocessing.} For Zhu-Net, CardiacSeg, and the SwinUNETR variants, volumes are cropped to a cardiac ROI using trained binary blood-pool localizers and z-score normalized over non-zero voxels. Zhu-Net and SwinUNETR preserve the native anisotropic image spacing prior to resizing, whereas CardiacSeg follows its provided configuration and resamples images to isotropic (1,1,1) mm spacing. All three methods then use $128^3$ inputs, with trilinear interpolation for images and nearest-neighbor interpolation for labels.

\noindent\textbf{nnU-Net configuration.} nnU-Net receives the ROI-cropped volumes without the fixed $128^3$ resizing applied to the other methods and independently configures its preprocessing, architecture, training, and inference pipeline for each training cohort. It selected anisotropic target spacings of $(0.60,0.525,0.509)$~mm for the combined 3D-Labs+HVSMR cohort and $(0.50,0.352,0.352)$~mm for the 3D-Labs-only cohort. In both settings, nnU-Net used batch size~2 and sampled $128^3$ training patches from the full-resolution resampled volumes, followed by sliding-window inference over the complete volumes.

\noindent\textbf{Self-supervised pretraining.} SwinUNETR variants pretrain a \texttt{SwinViT} backbone (\texttt{feature\_size}~$=48$, one input channel) with MAE or JEPA. MAE uses $75\%$ random masking and reconstructs masked voxel intensities; JEPA uses $40\%$ contiguous block masking and predicts EMA target features. CardiacSeg uses MAE-style pretraining of its ViT encoder with the same $75\%$ mask ratio as MAE-pretrained SwinUNETR, enabling a controlled comparison between masked-pretrained architectures. All SwinUNETR pretraining uses AdamW for 200 epochs, while CardiacSeg uses 400; with learning rate $1.5\times10^{-4}$, weight decay $0.15$, 10-epoch warmup, cosine decay to $10^{-6}$, and gradient clipping at $0.5$.

\noindent\textbf{Supervised training and fine-tuning.} nnU-Net uses its default combined Dice and cross-entropy loss, while Zhu-Net uses Dice loss. CardiacSeg and the MAE- and JEPA-pretrained SwinUNETR variants initialize their encoders from self-supervised pretraining and are fine-tuned end-to-end with no frozen layers, whereas the scratch SwinUNETR is randomly initialized; CardiacSeg uses its default combined Dice and cross-entropy loss, and SwinUNETR uses soft Dice loss. All fine-tuning uses AdamW with learning rate $1\times10^{-3}$, weight decay $10^{-4}$, warmup-cosine scheduling, and gradient clipping at $1.0$. Models are trained for up to 200 epochs with early stopping on validation Dice using patience 20.

\noindent\textbf{Repeated training and uncertainty reporting.}
Each model and training condition is evaluated across five independent seeds. Results are reported as mean $\pm$ standard deviation across runs, including the CHD-wise and structure-wise analyses.

\subsection{Results}

\begin{figure}[t]
    \centering
    \includegraphics[width=1\linewidth]{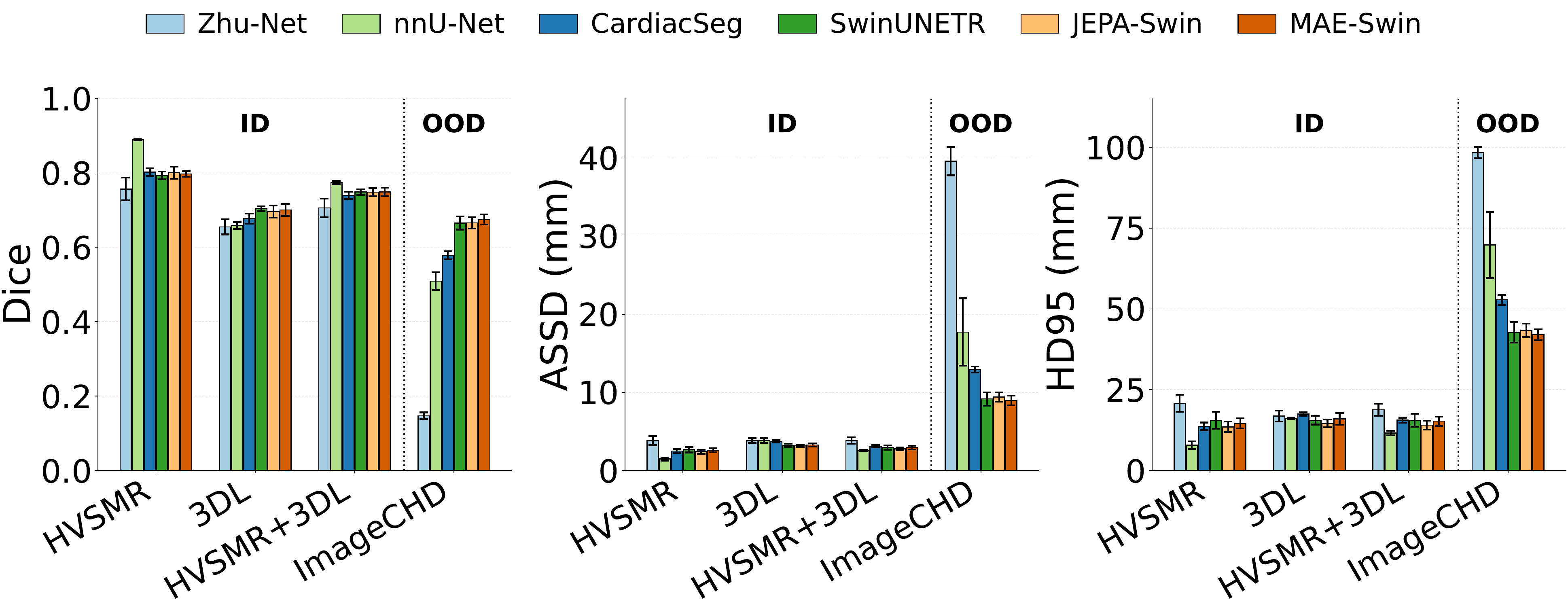}
    \caption{Segmentation performance for models trained on the combined CT+CMR cohort, evaluated on the in-distribution HVSMR and 3D-Labs validation splits and the out-of-distribution ImageCHD cohort. Bars show mean performance across five independent seeds, and error bars indicate $\pm$ one standard deviation.}
    \label{fig:combined_metrics}
\end{figure}

Figure~\ref{fig:combined_metrics} reports performance for models trained on the combined CT+CMR cohort, evaluated on both the in-distribution (ID) validation splits and the held-out ImageCHD set. Two patterns stand out. First, the ID and ImageCHD performance rankings differ substantially: nnU-Net attains the highest overall mean ID Dice of 0.77 but decreases to 0.51 on ImageCHD. Zhu-Net shows the weakest ImageCHD performance, reaching 0.15 Dice and the largest boundary errors among the evaluated methods (HD95 98.3~mm, ASSD 39.6~mm), while nnU-Net reaches an HD95 of 69.8~mm and an ASSD of 17.7~mm. Second, the SwinUNETR variants achieve the sbtrongest ImageCHD performance among the evaluated methods. SwinUNETR trained from scratch reaches 0.67 Dice with an HD95 of 42.7~mm and an ASSD of 9.2~mm, while MAE- and JEPA-pretrained SwinUNETR reach 0.68 and 0.67 Dice, respectively. Under combined CT+CMR training, MAE and JEPA pretraining therefore yield only marginal improvements over training SwinUNETR from scratch.

\begin{figure}[t]
    \centering
    \includegraphics[width=1\linewidth]{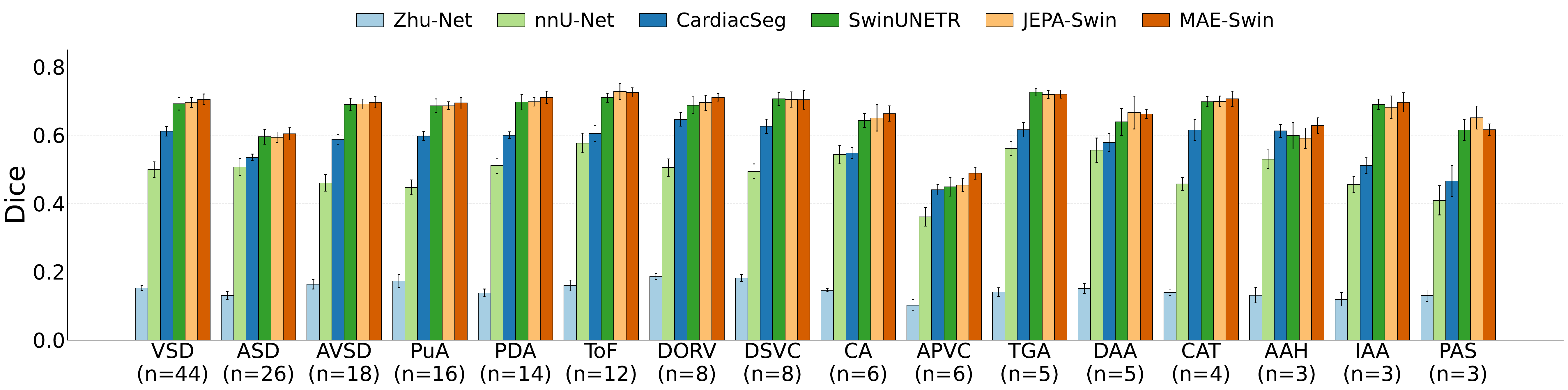}
    \caption{CHD-wise Dice performance on the ImageCHD cohort for models trained on the combined CT+CMR cohort. Bars show mean performance across five independent seeds, and error bars indicate $\pm$ one standard deviation.}
    \label{fig:chd_wise_ood}
\end{figure}

Figure~\ref{fig:chd_wise_ood} examines whether the aggregate ImageCHD results are consistent across CHD diagnoses. Performance varies across diagnostic groups, but the relative advantage of the SwinUNETR variants is maintained across most represented conditions. The lowest performance is observed for some of the least represented and anatomically complex groups, although these comparisons should be interpreted cautiously because several diagnoses contain only a small number of cases.

\begin{figure}[t]
    \centering
    \includegraphics[width=0.8\linewidth]{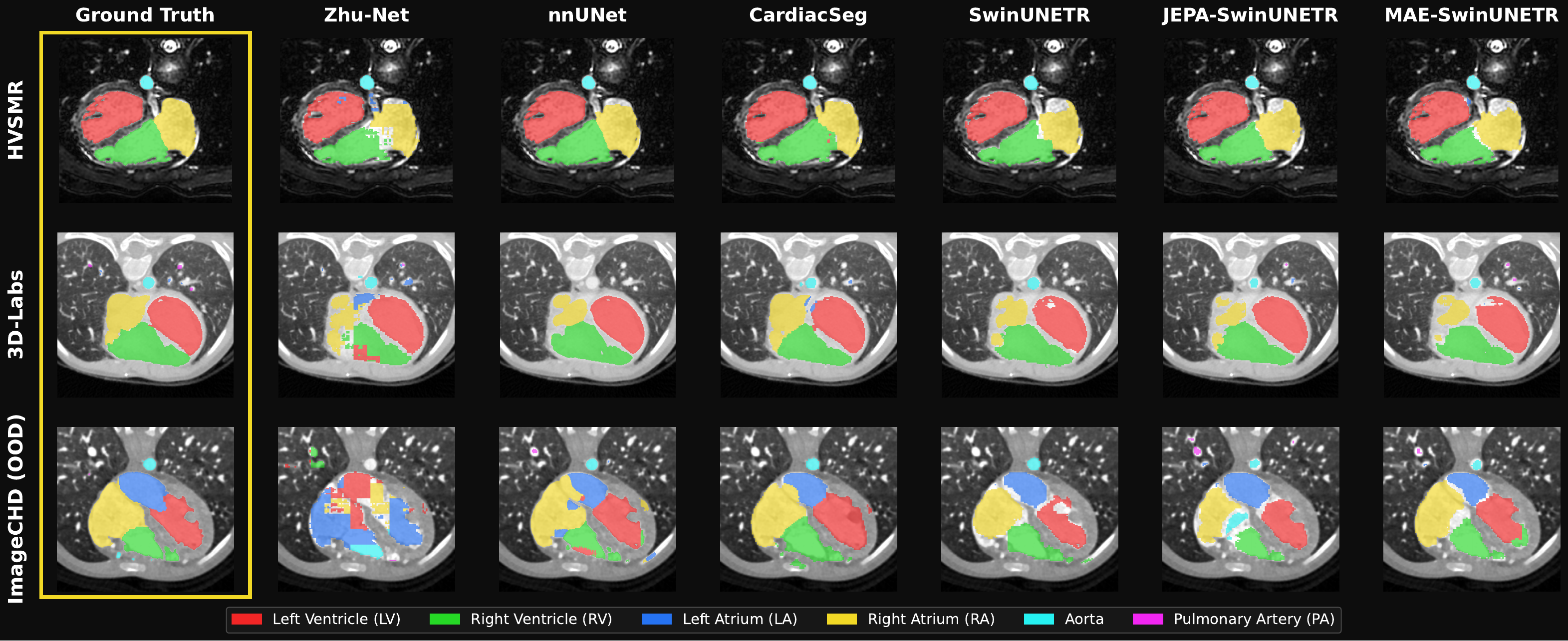}
    \caption{Qualitative comparison of segmentation predictions across the evaluated methods on representative in-distribution HVSMR and 3D-Labs cases and an out-of-distribution ImageCHD case.}
    \label{fig:qualitative_analysis}
\end{figure}

\noindent\textbf{Qualitative analysis.}
Figure~\ref{fig:qualitative_analysis} shows more coherent ImageCHD masks for the SwinUNETR variants, whereas Zhu-Net, CardiacSeg, and nnU-Net exhibit structure-level errors despite accurate predictions on the displayed in-distribution cases.

\subsubsection{Single-Modality Training: Private CT Cohort Only}
To measure the contribution of multi-modal training, we train the same architectures using only the private 3D-Labs CT cohort and evaluate OOD performance on ImageCHD. This setting removes HVSMR-2.0 CMR from training while preserving the same private CT training and validation split.

\begin{figure}[t]
    \centering
    \includegraphics[width=\linewidth]{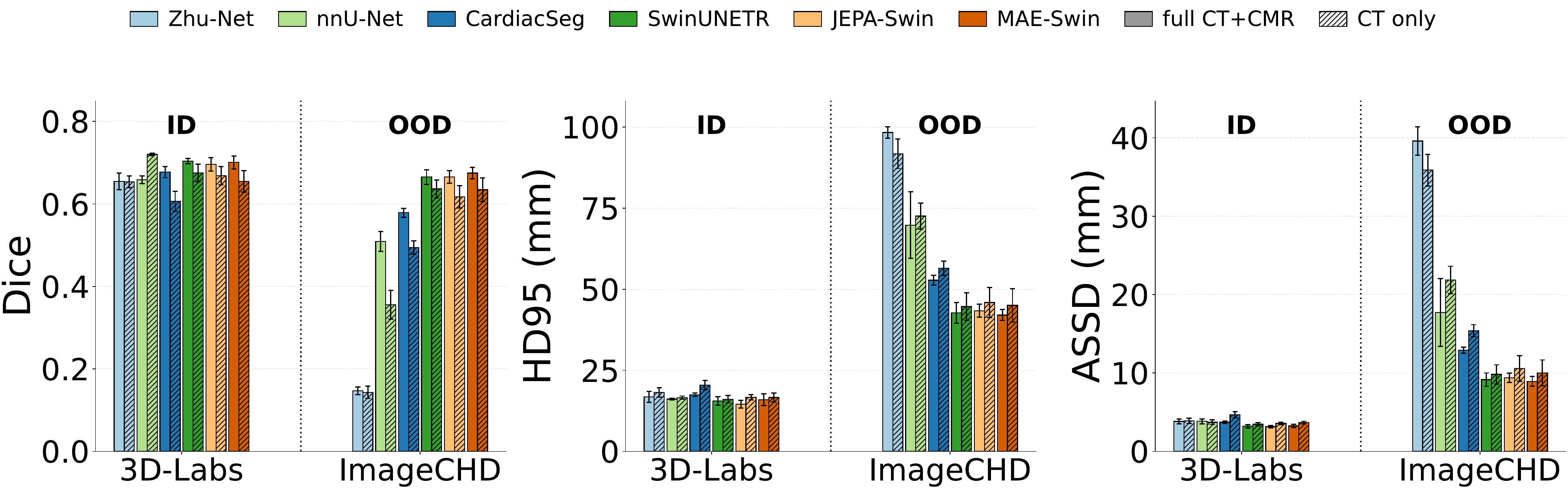}
    \caption{Effect of removing the HVSMR-2.0 CMR cohort. Bars show mean performance across five independent seeds, with error bars indicating $\pm$ one standard deviation. Solid bars denote combined CT+CMR training and hatched bars denote 3D-Labs-only training. Results are reported on the in-distribution 3D-Labs split and the out-of-distribution ImageCHD cohort.}
    \label{fig:ct_only_ablation}
\end{figure}

Figure~\ref{fig:ct_only_ablation} shows that removing HVSMR-2.0 reduces ImageCHD performance across the evaluated methods, but the magnitude of the effect differs considerably. SwinUNETR trained from scratch decreases from 0.67 to 0.64 Dice, while MAE- and JEPA-pretrained SwinUNETR decrease from 0.68 to 0.63 and from 0.67 to 0.62, respectively. CardiacSeg decreases from 0.58 to 0.49 and Zhu-Net from 0.15 to 0.14. The largest degradation occurs for nnU-Net, which decreases from 0.51 to 0.36 Dice when the CMR cohort is removed.

Thus, among the evaluated methods, SwinUNETR retains the strongest ImageCHD performance when training is restricted to the private CT cohort. More broadly, the large variation in performance degradation after removing HVSMR-2.0 shows that the contribution of joint CT/CMR training is strongly method-dependent. However, these comparisons do not isolate architecture from other differences between the evaluated models and training pipelines.

\noindent\textbf{Structure-wise performance.}
Table~\ref{tab:structure_dice} reports ImageCHD Dice separately for the six anatomical structures under combined CT+CMR and 3D-Labs-only training. The results reveal substantial variation across anatomical structures and complement the aggregate evaluation in Figs.~\ref{fig:combined_metrics} and~\ref{fig:ct_only_ablation}.

\begin{table}[tb]\centering
  \caption{Per-structure Dice on the ImageCHD OOD cohort (mean $\pm$ SD across five seeds). Overall denotes the mean across the six anatomical structures.}
  \label{tab:structure_dice}\footnotesize\setlength{\tabcolsep}{3.0pt}
  \begin{tabular}{@{}lccccccc@{}}\toprule
  Method & LV & RV & LA & RA & Ao & PA & Overall \\ \midrule
  \multicolumn{8}{@{}l}{\textit{Combined CT+CMR training}} \\[1pt]
  Zhu-Net    & $.08 \pm .05$ & $.02 \pm .01$ & $.06 \pm .02$ & $.14 \pm .05$ & $.37 \pm .02$ & $.21 \pm .04$ & $.15$ \\
  nnU-Net    & $.57 \pm .02$ & $.38 \pm .04$ & $.54 \pm .04$ & $.53 \pm .02$ & $.61 \pm .15$ & $.42 \pm .03$ & $.51$ \\
  CardiacSeg & $.58 \pm .02$ & $.55 \pm .02$ & $.65 \pm .01$ & $.63 \pm .01$ & $.62 \pm .00$ & $.44 \pm .02$ & $.58$ \\
  SwinUNETR  & $\mathbf{.70} \pm .03$ & $.63 \pm .02$ & $.74 \pm .02$ & $.70 \pm .04$ & $.68 \pm .01$ & $.54 \pm .03$ & $.67$ \\
  MAE-Swin   & $\mathbf{.70} \pm .02$ & $\mathbf{.64} \pm .02$ & $\mathbf{.75} \pm .02$ & $\mathbf{.72} \pm .03$ & $\mathbf{.69} \pm .03$ & $\mathbf{.56} \pm .01$ & $\mathbf{.68}$ \\
  JEPA-Swin  & $.67 \pm .02$ & $\mathbf{.64} \pm .03$ & $\mathbf{.75} \pm .01$ & $\mathbf{.72} \pm .03$ & $.68 \pm .03$ & $.55 \pm .01$ & $.67$ \\
  \addlinespace[3pt]
  \multicolumn{8}{@{}l}{\textit{3D-Labs-only training}} \\[1pt]
  Zhu-Net    & $.07 \pm .04$ & $.03 \pm .03$ & $.08 \pm .03$ & $.10 \pm .01$ & $.34 \pm .04$ & $.24 \pm .05$ & $.14$ \\
  nnU-Net    & $.32 \pm .05$ & $.29 \pm .07$ & $.43 \pm .02$ & $.39 \pm .04$ & $.38 \pm .14$ & $.32 \pm .06$ & $.36$ \\
  CardiacSeg & $.43 \pm .03$ & $.48 \pm .04$ & $.59 \pm .02$ & $.54 \pm .02$ & $.55 \pm .01$ & $.38 \pm .02$ & $.49$ \\
  SwinUNETR  & $.66 \pm .02$ & $\mathbf{.58} \pm .02$ & $.70 \pm .05$ & $\mathbf{.70} \pm .04$ & $\mathbf{.65} \pm .02$ & $\mathbf{.53} \pm .01$ & $\mathbf{.64}$ \\
  MAE-Swin   & $\mathbf{.67} \pm .03$ & $.57 \pm .04$ & $\mathbf{.72} \pm .02$ & $.67 \pm .07$ & $\mathbf{.65} \pm .02$ & $.52 \pm .04$ & $.63$ \\
  JEPA-Swin  & $.63 \pm .02$ & $.53 \pm .05$ & $.70 \pm .05$ & $.69 \pm .02$ & $.64 \pm .02$ & $.50 \pm .05$ & $.62$ \\
  \bottomrule
  \end{tabular}
\end{table}

\subsection{Target-Domain Sample Efficiency}
To quantify target-domain sample efficiency, we progressively add labeled ImageCHD cases to training in increments of 11 and evaluate on a fixed 22-case ImageCHD holdout.

\begin{figure}
    \centering
    \includegraphics[width=1\linewidth]{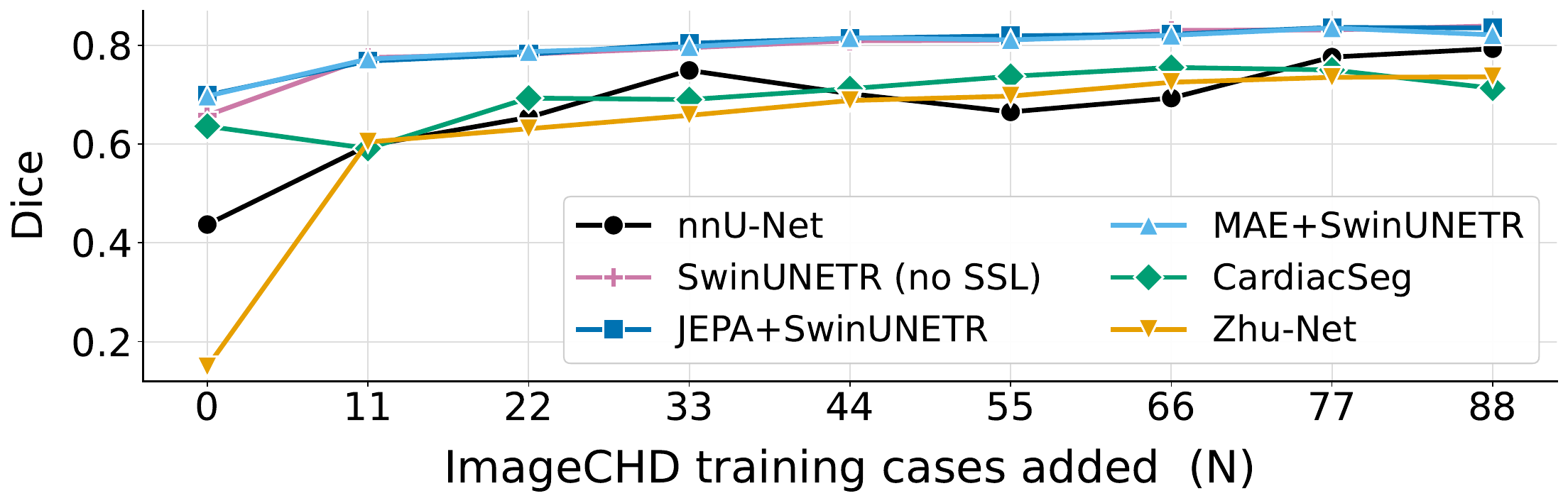}
    \caption{Target-domain sample-efficiency analysis on ImageCHD. Labeled ImageCHD cases are added in increments of 11 while performance is evaluated on a fixed 22-case OOD holdout ($N{=}0$ is zero-shot).}
    \label{fig:ablation_study}
\end{figure}

Figure~\ref{fig:ablation_study} shows that the SwinUNETR variants provide the strongest low-label performance among the evaluated methods on the fixed ImageCHD holdout. The MAE- and JEPA-pretrained variants achieve the highest zero-shot Dice, and all SwinUNETR variants exceed $0.76$ Dice after adding only 11 labeled ImageCHD cases. Zhu-Net and nnU-Net show large gains after the first set of target-domain labels but do not consistently match the SwinUNETR variants across the evaluated sample range. CardiacSeg improves more gradually and approaches the SwinUNETR variants at higher label counts but does not surpass the best-performing variant. Within this ImageCHD adaptation experiment, SwinUNETR therefore provides the strongest low-label performance, while the benefit of self-supervised pretraining is concentrated primarily in the zero-shot regime.

Taken together, the combined-cohort, CT-only, and target-domain adaptation experiments identify the SwinUNETR variants as the strongest evaluated methods across the studied ImageCHD settings. SwinUNETR trained from scratch remains competitive with its MAE- and JEPA-pretrained variants, showing that the tested self-supervised initializations provide only limited additional gains. These results do not establish architecture as the causal source of the observed performance differences or demonstrate that the same model ranking will hold across other OOD cohorts. 

\subsection{Interpreting the SwinUNETR Effect}

SwinUNETR's stronger ImageCHD performance may stem from its hierarchical shifted-window Transformer encoder, which integrates local and cross-window spatial context, together with a convolutional decoder that preserves anatomical detail. MAE and JEPA provide only modest gains, likely because pretraining is limited to the same relatively small source cohort without external unlabelled data; their advantage is most evident in the zero-shot regime and diminishes once target-domain labels are introduced.

\subsection{Limitations}

ImageCHD is dominated by septal defects such as VSD and ASD, whereas the private source cohort contains a greater proportion of complex CHDs, including DORV; the findings therefore reflect this specific source-target shift. For methods using the common pipeline, resizing volumes to $128^3$ may reduce fine anatomical detail, while broader hyperparameter tuning could further improve performance. The observed advantage of SwinUNETR should therefore be interpreted within the evaluated datasets and experimental settings.

\FloatBarrier
\section{Conclusion}
We present a systematic OOD evaluation of CHD segmentation across representative models, multi-modal training settings, and self-supervised pretraining strategies, using ImageCHD as a held-out target cohort. Across the evaluated methods, in-distribution validation ranking does not align with performance on ImageCHD: nnU-Net achieves the highest combined validation Dice but decreases substantially on the held-out cohort. The SwinUNETR variants achieve the strongest ImageCHD performance and target-domain sample efficiency, including when SwinUNETR is trained from scratch. MAE and JEPA pretraining provide only modest additional gains under combined CT+CMR training and no improvement over scratch SwinUNETR in the CT-only experiment. Joint CT/CMR training improves ImageCHD transfer for most of the evaluated methods, although the magnitude of this effect differs considerably between models. These findings demonstrate the importance of complementing in-distribution validation with explicit cross-dataset evaluation and identify SwinUNETR as the strongest evaluated model for the studied transfer to ImageCHD. Evaluation on additional target cohorts and more controlled comparisons are required before attributing these findings conclusively to architecture or extending them to CHD segmentation under distribution shift more broadly.

\paragraph{Acknowledgments.}
This work was supported by the Patrick J McGovern Foundation.

\paragraph{Disclosure of Interests.}
The authors have no competing interests to declare that are relevant to the content of this article.

\FloatBarrier
\bibliographystyle{splncs04}
\bibliography{references}
\end{document}